\documentclass[11pt]{article}
\usepackage{acl2016}
\usepackage{times}
\usepackage{url}
\usepackage{latexsym}

\aclfinalcopy 

\let\OLDthebibliography\thebibliography
\renewcommand\thebibliography[1]{
  \OLDthebibliography{#1}
  \setlength{\parskip}{4.5pt}
  \setlength{\itemsep}{0pt plus 0.3ex}
}

\title{Throwing fuel on the embers: \\ Probability or Dichotomy, Cognitive or Linguistic?}

\author{David M. W. Powers \\
  Artificial Intelligence and Cognitive Science Group \\
  Flinders University \\
  Adelaide, South Australia \\
  {\tt David.Powers@flinders.edu.au} \\
}

\date{}

\begin{document}
\maketitle
\begin{abstract}
 Prof.\ Robert Berwick's abstract for his forthcoming invited talk at the ACL2016 workshop on Cognitive Aspects of Computational Language Learning revives an ancient debate. Entitled ``Why take a chance?", Berwick seems to refer implicitly to Chomsky's critique of the statistical approach of Harris as well as the currently dominant paradigms in CoNLL. 

Berwick avoids Chomsky's use of ``innate" but states that ``the debate over the existence of sophisticated mental grammars was settled with Chomsky's Logical Structure of Linguistic Theory (1957/1975)", acknowledging that ``this debate has often been revived".  Berwick does not in his abstract mention Gold's concerns about what it means to learn a language, or his concept of ``Language Identification in the Limit" which is often juxtaposed against the idea of ``Probably Approximately Correct", but Berwick specifically questions the counterargument that ``differences in acceptability judgements can be closely approximated ... from probabilistic accounts". We discuss his recent critique of computational experiments that show  acceptability judgements correlating with probabilities from simple computational models.

This paper agrees with the view that this debate has long since been settled, but with the opposite outcome!  Given the embers have not yet died away, and the questions remain fundamental, perhaps it is appropriate to refuel the debate, so I would like to join Bob in throwing fuel on this fire!  
\end{abstract}

\section{Introduction}

In a recent paper at GLOW16 and an invited talk to be given at CACLL, Berwick et al.\ (2016ab) make the strong claim that ``debate of the existence of sophisticated mental grammars was settled with Chomsky's The Logical Structure of Linguistic Theory (1957/1975)... But this debate has often been revived." Yet if it has been ``settled" that begs the question of why it keeps being ``revived".  This conundrum reflects different views of the nature of science, and the nature of learning and cognition, as much as the nature of language.

The reason for these divergent views of the history of the theory of language learning may relate to the shifting ground on which the battle has been fought.  In fact, it is not even clear whether the focus of the debate is what is innate, what is modular, what is universal, what is recursive, ``gradience" of grammaticality, or validity of theories captured in terms of three letter acronyms: LAD, PoS, TGG, EST, P\&P, G\&B, etc. 

But then there is also Chomsky's own (1986/1991/1993/1995) ``Minimalist Program" (MP).  Chomsky doesn't regard things as settled by his 1957-75 work and specifically critiques it. Of further concern is the frequent explanation that ``evolution did it" or, more tongue-in-cheek, ``God does it on Tuesdays"(Fodor in Piatelli-Palmarini, 1979). For Computer Scientists and Computational Linguists, evolutionary and genetic methods are  weak members of a broader set of Machine Learning algorithms.  Claiming something cannot be learned with the full power of the machine learning arsenal, but has been learned with just the weaker evolutionary techniques, is oxymoronic. Alternatively it can be viewed as contradicting either the Church-Turing thesis or the assumption that the human brain is some kind of machine. 

\section{Poverty of Science}

While Pinker (1984, 1994, 1997) adheres to the Chomskian paradigm, defending an innate specialized modular account of our language capacity, he treats our scientific capacity within the same framework, thus making both our language and science evolutionary artefacts rather than ways to model the world. This blurs the distinction between folk biology/psychology/theory and science unless by applying this limited rationality we have managed to develop sound logics for argumentation, as perhaps expressed in formal mathematics, learning algorithms, and falsificationist accounts of science. Carruthers et al.\ (2002/2004) set out to explore the Pinker account and discount his negative conclusions about science. What distinguishes folk science from real science is that there is a logic of proof that requires evidence, while evidence that led to formation of our hypotheses, even unconsciously, or training data that was used to learn a model, even indirectly, cannot be used as evidence in support of a theory. This is particularly a problem for evolutionary theory.

Nagel (2012, p4)  states: ``I would like to defend the untutored reaction of incredulity to the reductionist neo-Darwinian account of the origin and evolution of life... What is lacking... is a credible argument that the story has a nonnegligible probability of being true." Ironically some proponents of evolutionary accounts seem religious in reliance on prior beliefs and neglect of their status as assumptions or hypotheses, and Nagel seeks to find a dualist middle between religious extremes. 

The issue is forgoing the scientific discipline of providing refutable predictions of the theory (Popper,1934/1959; Lakatos,1970,1976,1978ab).  That is, an evolutionary just-so ``story" (Nagel,2012) or ``myth" (Popper,1963)  is not sufficient in ``materialist naturalism" or ``reductionism" or ``empiricism" (Nagel,2012,p13f; Popper,1934,1965), the basis of the modern scientific revolution that goes back to Galileo and Descartes and explicitly leaves out appeals to human perception and intention (Nagel,2012,p35). Indeed Chomsky (1995) now avoids the word ``theory" for his Minimalist Program.  Use of the word ``program" following Lakatos (1970) implies a base of ``minimal" ``hard core" assumptions that are assumed but \em not \em subject to falsification, which are assumed along with further auxiliary hypotheses that \em are \em subject to change in response to empirical evidence.

\section{Poverty of Evolution}

It is clearly difficult to make provable predictions for evolutionary theories - Darwin's theory was an example of \em pseudoscience \em in Lakatos (1970). But there are ways, and in particular we can adopt the \em computational metaphor \em and treat \em evolution as an algorithm \em. To provide support for LAD versus its alternatives, its predictions need to differ from those that derive from alternate models of language acquisition, development and emergence, such as those provided by Cognitive Linguistics. A Computational Cognitive Linguistics model proposes mechanisms based on ideas of similarity (metaphor) within and across modalities (Powers,1997) that are important for survival (fitness) and which also predict other higher level phenomena (analogy, metaphor, metonymy, etc.), and provide accounts for superstitions and illusions. Chomsky (Piatelli-Palmarini, M., 1979) refused to be drawn on how ``evolution" achieved his LAD, saying it was not his job as a linguist, which is divorcing linguistics from science. 

Computational Intelligence split off historically from Artificial Intelligence over inclusion of such `non-logical' algorithms in our arsenal, although today evolutionary and genetic techniques, as well as others modeled on  colonies, swarms, etc. are used regularly across individuals in combination with adaptive learning within the individual to try to develop the appropriate structure for behavioural learning that will \em maximize the fitness \em of the individual in the environment. Often this is done without forcing a strict non-communication of learned information to descendants as enshrined in both pre-genetic evolutionary theory (life before genes) and neo-Darwinian Mendelian evolutionary theory (strictly speaking genetic models). 

While excluded by Cartesian science (Nagel, 2012), Beliefs, Desires and Intentions (BDI) are specifically brought back in by a dominant approach to reasoning in Artificial Intelligence. Questions of consciousness are also explored both in Artificial Intelligence and Cognitive Science, and from the perspectives of language and learning, the ``acceptability judgements" that Chomsky (1957) and Berwick (2016ab) appeal to depend on hidden variables (normally handled preconsciously) being brought to conscious attention - with the result that acceptability cannot be explained without appeal to theory, hypotheses and assumptions, and is self-reinforcing and biased. 

\section{Poverty of Dichotomy}

This begs the question of whether issues with acceptability are syntactic, semantic or pragmatic, and Berwick et al. (2016a) expect them to be syntactic and dichotomous rather than the gradient of gramaticality that Lau, Clark \& Lappin (LCL:2014,2015) explore by correlation of human Likert-scale judgements with a statistical model. Berwick questions whether observed correlations are strong enough to warrant accepting that acceptability was a matter of probability (wrapping up a question about gradience of human judgements with a question about the probablistic nature of grammar). The fact that particular simplistic probabilistic or neural models don't correlate as well as would be liked with human judgements, with no competing LAD-PAP model constrasted with it, proves nothing about whether LAD-PAP is correct, language is probabilistic, or acceptability is graded (nor does it refute it clearly, given they suggest other causes for the small correlations seen). The fact that some correlation is observed in both the original and the replication study, provides some support for the graded probabilistic account as it is a prediction of the theory, but is not conclusive in that direction either.

Berwick et al.\ (2016a) use data from the generative linguistics tradition, from papers in Linguistic Inquiry, two sets of examples of acceptable and unacceptable sentences from textbooks, plus 120 permutations of Chomsky's ``Colorless green ideas sleep furiously" (inspiring the title of the replication paper).  LCL (2013-15) used examples from the British National Corpus (BNC) with ``infelicities introduced by round trip machine translation". The Berwick results showed correlation, but weaker than in the original experiment. But, real sentences from a corpus of real English have a Zipfian distribution, can be well over 100 words long, and will thus tend to be much longer and more complicated than short textbook-style examples composed to make a particular point (and LCL made a point of examining length and normalizing for it). Also machine translation into English will produce a different distribution of ``infelicities". Finally, much of the `starring' of sentences in linguistic publications, particularly textbooks, is naive, while LCL dealt with a corpus of real English and occasionally focussed in on one of the most prevalent issues of ambiguity (past tense versus passive participle). 

\section{Poverty of Tagging}

The most common reason for \em mis\em starring sentences is part of speech (POS) ambiguity, not recognizing the flexibility of English to coerce a word that is intrinsically or overwhelmingly one POS: a meaning may only be licenced in specific contexts so that the grammatical possibility is not salient otherwise. Some words have an auxiliary or modal role and a literal meaning (be, have, go, ...). A noun can in general be used to express an action involving that object (e.g. a body part - ``He \em shoulders \em the player aside, \em heads \em the ball to a team mate, then \em legs \em it back to the goal."). This is explored by Entwisle  (1997, 1998), leading to a critique of parsers and metrics that `improve' by ignoring such choices and tagging with, or biasing to, the most frequent POS. Parsers trained and tested on handtagged subsets of a corpus then tag the whole corpus leading to unreliable tags, and even after attempts to patch them still have cases that fail systematically. BNC is used in the LSL (2013-15) studies, but its tags were ignored. They still exhibit problems  (particularly when participles are involved, or words with concrete and abstract or functional usages). Of 278 instances of \em going to work \em in BNC, only 4 are ambiguous when viewed in context, and of the remaining 274, we find 107 are correctly tagged (78 of 88 nouns and 29 of 186 verbs) and 167 are incorrectly tagged (10 of 88 nouns and 157 of 186 verbs). This performance is not significantly different from chance ($p>0.1$), with the probability of an informed tag for ``work" being 4.4\% (Powers,2008, 2012).

The conclusions drawn by Berwick are unrealistic in expecting that an unsupervised word-level trigram model, or a simple RNN, should provide high performance: it is known that at each level of linguistic processing, both forward and backward context of two or more units is indicated as context necessary for correct interpretation (phonological/phonemic, morphological/semantic, grammatical/syntactic, ontological/pragmatic, etc.) and this multilevel multiunit multimodal information is routinely used in speech recognition, machine translation, parsing, etc. and higher order modeling has been shown to lead to better entropy estimates and evaluations. LCL have the opposite perspective, claiming only that simplistic models like theirs have predictive ability on real sentences, compared with Berwick et al.'s (2016a) datasets that look like teaching examples.

\section{Poverty of Psychology/Biology}

Chomsky is known for his political and philosophical arguments, and polemical style, as much as for his linguistic contributions, and his arguments in his linguistic writing often share this political, polemical and philosophical flavour.  His debating style can in many places be characterized as ``intentionally obtuse for the purpose of scoring a debating point" (Palmer,2006, p255; Piatelli-Palmarini, 1979). Indeed such discussions of \em -isms \em characterize the butting and abutting of disciplinary viewpoints that triggered the emergence of Cognitive Science generally, as well as Cognitive Linguistics specifically, as arguments spilled across disciplinary lines and researchers tried to synthesize interdisciplinary theories that were consistent across different forms of theory and evidence.  Chomsky's (1959) ``linguist" perspective on Skinner's (1957) ``behaviourist" treatment of language was in fact what really catapulted him to fame, although again here he was faulted both for his lack of understanding of ``behaviourism" and his ``several serious errors of logic" (MacCorquodale,1970)  and ``revealed misunderstandings so great that they undercut the credibility of the review substantially... a kind of ill-conceived dam in the progress of science, a rhetorically effective but conceptually flawed document that would eventually be overborne" (Palmer,2006,p253; 2000).

Chomsky (1957) had no biological evidence that specific hypothesized \em parts of speech \em (Noun, Verb, etc.) and \em grammatical constructs \em (word, phrase, clause, sentence, etc.) or \em linguistic modules \em (phonology, morphology, etc.) actually exist in the mind of the native speaker in any sense. Even today, when modern neuroscience is starting to be able to look at what is happening during language processes, the maps do not correspond straightforwardly to specific linguistic predictions or theories, or show loci for words or POS (Pulvermuller,1999; Huth et al.,2016). However, we can do better now than just talk about Broca's area being near the motor areas presumably involved in producing speech and Wernicke's area being near the auditory areas presumably involved in understanding speech, with the observation that the disruption caused by damage to these areas affects more than just the nearby modality gives insights and a more sensorimotor foundation for a theory of language (Powers and Turk, 1989).

\section{Poverty of Generative Linguistics}

Chomsky is also attributed with bringing `generative' formalism to Linguistics, but as Pullum (2011) points out Chomsky's first published book ``Syntactic Structures" (SS: Chomsky (1957) did not address the question of whether English was a Finite State language, or propose  an innate Language Acquisition Device (LAD) in the 1950s, and nor did he in ``The Logical Structure of Linguistic Theory" (LSLT) the draft typescript from 1955, updated in manuscipt in 1956, and cited by Berwick in the form in limited circulation in the US in 1957, nor even in the final severely revised form published in 1975. SS (Chomsky, 1957) does advocate Transformational Generative Grammar (TGG), building on an unformalized notion of transforms that Harris is suggested to have borrowed from Carnap, and which is used inconsistently giving incorrect results in SS with a treatment \em in\em compatible with LSLT (Pullum, 2011) . 

The mathematical formalism behind the TGG, and indeed the so-called Chomsky hierarchy, goes back to Post, who in turn drew on Whitehead and Russel (Pullum, 2011). By the time LSLT (Chomsky, 1957/75) was published, the comprehensive review of Heny (1979) notes, Chomsky was already withdrawing his support from TGG and had reduced his number of proposed tranformations down to one or two. Eventually in the Minimalist Program (Chomsky, 1995), the distinction between deep structure and surface structure had been eliminated in favour of a more explicitly derivational approach, with X-bar theory devolving to pure phrase structure! Thus it is hard to see that anything was ``settled" by LSLT in 1957.

Anderson (2007/8) makes LSLT the title of a Presidential Address to LSA, and revisits many of these issues - hardly ``settled". Every year this debate has been ``revived" in multiple papers and books and venues like Behavioural and Brain Sciences. It forced the founding of the Cognitive Linguistics Journal and associated societies: rather than generated from an internal LAD the CogL focus is on the metaphor-driven emergence of language from general mechanisms for understanding and interacting with our world (Lakoff and Johnson,1980,1987). Powers (1983,1991b,1997) connects this to the Psycholinguistic theory of Piaget (1923, 1936, 1950; Piatelli-Palmarini,1979) and the generalized Emic theory of Pike (1947, 1948,1954, 1955). 

\section{Poverty of Grammar}

Powers and Turk's (1989) argument that Chomsky's approach is not neurologically or computationally plausible presents evidence of violating Chomsky's agreed assumptions (Hockett,1961), as well as discussing the implications of computational language learning models. We have a rich supply, with early models including Adriaans (1992ab,1993), Anderson (1975), Block (1975), Ellison(1993), Entwisle (1997), Finch (1993), Gold(1967), Horning(1969), Powers (1983, 1991a), Siklossy (1971) and Turk (1984,1988,1990) and extending beyond the reaches of grammar! It is  worth noting that the phonology work of Bird and Ellison (1994) traces back to that of Goldsmith (1976) who framed his work (with some difficulty it seems) in the Chomskian framework, while the semantic directions relate specifically to the idea that language is impossible without what Hayes (1979) calls Naive Physics, Powers (1983, 1991ab) calls Ontology,  Harnad (1990,1991) calls Symbol Grounding, or Feldman, Lakoff, Stolcke \& Weber (1990) call L0. 

This goes beyond dictionaries, thesauri, semantic nets and taxonomies to require multimodal neural connections between multiple sensory-motor modalities and thus establish real meaning in the physical world rather than just chasing words around a dictionary or its electronic equivalent. It provides an answer for debates around Searle's (1980) Chinese Room and Harnad's (1990) Total Turing Test, and the broader Connectionism vs Symbolism Debate (Powers,1992), and acknowledges the supervisory possibilities of multimodal learning (Powers and Turk, 1989) and memory-based anticipated correction (Turk, 1984, 1988, 1990).  This also connects to Berwick and Chomsky's focus on so-called \em generative \em grammar - in fact, Powers and Turk (1989) discuss evidence of \em separate \em learning of language \em understanding \em models near \em sensory/hearing \em processing areas of the brain (e.g. Wernicke's area) and language \em generation \em models near \em motor/speech \em centres of the brain (Broca's area), as well as prediction of \em mirroring \em regions that connect what we hear and see with what we produce ourselves and others reflect (mothers mirror their babies more than vice versa). 

This can be seen to predict the recent discovery of mirror neurons (Arbib, 2009), and builds on classic Piagetian treatment of \em reflection \em and \em imitation \em in learning (Piaget, 1923,1936,1950). 

\section{Poverty of the Stimulus}

Mirror neurons fire not only when the subject performs an action, but when they see someone else perform the action. We subdivide imitation into three distinct paradigms: 2a. the child speaks and the parent echoes (mirroring), 2b. the parent speaks and the child echoes (echolalia), and 2c. the child speaks and is directly mirrored (via a physical mirror or a camera and monitor setup). These are listed in decreasing order of importance and commonality: the parent echoes the child much more than vice-versa (Oostenbroek et al.,2016), and this is enshrined in the kind of peek-a-boo games played with infants, as well as the encouragement and shaping of the child's protowords (‘Dada – she said Dada!), on the other hand the mirror or camera is not usually available, and the child must actually learn to recognize herself, and the fascination with mirrors and cameras thus comes at a relatively late stage.

In the end the real problem with the Nativist and the Symbolist positions is that they rely only on explicit linguistic `stimulus' - humans rely on connecting symbols with real world sensory input and motor output, as well as proprioceptive and mirrored signals that combine and relate sensory and motor information. 

The Poverty of the Stimulus (PoS) argument focuses on syntax to the exclusion of semantics and pragmatics, experience and ontology. On the other hand Cognitive Linguistics emphasizes the role of similarity or metaphor in language, seeing patterns learned in the real world as models for those learned in language (e.g. Lakoff et al.,1980,1987; Goldberg,2006,2008), while Computational Intelligence emphasizes the role of similarity or correlation in learning so that multimodal similarity should be fundamental to Computational Cognitive Linguistic models of language learning and be consistent with physical limitations on memory and processing as well as the computational limitations on language that result, with (Powers,1997) proposing a variant blackboard model embedded in neural strata (shifting through the layers in time, with versions of grammar rules implicitly enshrined in each layer). Bod (as recently as 2009), commenting on Goldberg (2006,2008), decried the lack of a learning model, of a connection between Natural Language Learning and Cognitive Linguistics, emphasizing that it is important to make one.

\section{Poverty of the Poverty argument}

Chomskian linguistics relies heavily on Gold (1967) in arguing for Poverty of the Stimulus (PoS), asserting that language can't be learned because babies do not receive negative information (stimulus) required for learning. 

However, the results of Gold (1967), Horning (1969) and Adriaans (1992ab) that superfinite languages can be learned without such explicit stimulus are ignored. Gold (1967) noted that his proof \em broke down \em if the input was ordered, for example if a simple usage of a construction was used before a recursive usage of the construction (calling this ``anomalous text" and opening the door to ignoring it). Horning (1969) generalized this to probabilistic information (viz. seeing clear non-recursive constructs sufficiently often to overcome the noise-like effects of overly complex constructs, consistent with psycholinguistic evidence that ``we can only learn what we almost already know"). Adriaans (1992ab) extended to a categorial grammar, showing that language learning from a corpus was possible in this context. 

Based on psycholinguistic evidence, Powers and Turk (1989)  argue that ``a child only learns what they almost already know" , that the criterion of ``anomalous test" can be met by a self-organized filter, and that the ``anticipated correction" is provided  unsupervised (Turk,1984,1988,1990). This is the idea of ``acceptability'' applied to one's own speech! This is what a native speaker/learner would accept as normal sounding speech, and mixes up accent, phonetic exactitude, aptness of word choice, and other semantic and pragmatic aspects, with the pure syntactic idea of (grammatical) acceptability that Linguists employ, and illustrate with made-up sentences in text books.

Thus measuring acceptability based on written sentences is measuring acceptability in a language that is technically different from the `evolved' Natural Language of untutored speech, as it is a `prescriptive' language that is taught in schools, and we also show that it rises above the computability limitations on the complexity of natural speech that are imposed by the size of the human brain. Experiments with complex (e.g. centre-embedded) constructs do demonstrate that a higher complexity can be dealt with in writing compared with speech (and often the inability to deal with a construct is marked as unacceptability by the non-linguist).

\section{Poverty of the Acceptability argument}

It is important to observe that the Chomskian assumption that language is recursive is just an assumption (if part of the Minimalist Program). In terms of Theory of Computation, it is arguable that language is recurrent rather than recursive - that is there is a finite amount of real estate that is reallocated as needed, \em accepting \em sentences with a non-contracting Context Sensitive Grammar (CSG) rather than an unbounded `stack' that grows as needed, \em generating \em sentences of unbounded length and complexity. 

Due to its finite size, the brain is a Finite State Automaton (FSA) not a Pushdown Automaton (PDA) or Turing Machine (TM). A \em generative \em Context Free Grammar (CFG) requires an \em unbounded \em stack to \em generate \em its full range of sentences, while a Regular Grammar (or FSA) is equivalent to iteration or recurrence and doesn't require an unbounded stack to \em generate \em sentences with full range complexity. Powers (1997) embedded a cycling blackboard model of cognition into a finite network where sensorimotor stimuli and responses are processed through layers of neurons. Analogous processing takes place over multiple layers and modalities, and the same CFG or CSG rule can be \em reflected \em while \em disallowing \em recursion.

The invention of \em writing\em, adds a stack of paper (so person+paper = PDA). Thus written language can exceed the bounds on spoken language and potentially prove the reality of recursion (except that those schooled in Reading, wRiting and aRithmetic have been schooled in the basics of prescriptive grammar, and thus \em should \em be using Recursive Rules they have been schooled in, rather than \em Natural Language \em - this effect of teaching of invented grammar rules is seen in the mangled constructions students, and teachers of English, come out with as they try to speak proper by applying the rules they paid for in their education).

The invention of the \em eraser\em, and the ability to go back and forward in our notes \em without \em destroying them, provides us with an essentially unlimited amount of Turing Machine tape we can move back and forwards on. More conveniently, today we use word-processors, with essentially unbounded file space, which allows us to go back and change what we have written. A person with a word processor, or pen+eraser+paper, is a Turing Machine.  The Church-Turing thesis says there is nothing more powerful.

\section{Conclusion}

Here we have treated in each column issues that are debated in other places at booklength, reviewing less accessible pre-web theses and books plus early work that ``settles" \em against \em Chomsky (1957/75), as well as considering the recent contributions replicated with TGG examples by  (Berwick et al., 2016a). Now we focus concluding remarks on Berwick's titles, emphasizing that the debate is anything but settled \em by \em Chomsky.
 
\subsection*{``Why take a chance?''}

Acceptability judgements are influenced by contextual factors (the PoS examples) as well as modality (spoken vs written) and of course register, dialect, idiolect, etc.\ (for the purposes of this paper we go along with the \em assumption \em that there is such a language as ``English"). With acceptability judgements we balance comprehensibility (I can figure out what was meant) with aptness (it could have been better expressed) with innovation (a new metaphor, metonymy, personification, figure of speech, or PoS-generalization has been used) with actual clear syntactic error - though their origin may be typos or substitutions not competence (Huang and Powers,2001; Powers,1997).

Far from settling a debate, Chomsky's The Logical Structure of Linguistic Theory (1957/75) initiated a furore with an array of \em assumptions \em and \em assertions \em that many have thought dead and gone, including many that were discarded in Chomsky's Minimalist Program (1986/1991/1993/1995). If anything, since the emergence of Cognitive Science, Computational Cognitive Science, Cognitive Linguistics, Computational Psycholinguistics and their permutations,  the various interdisciplinary communities have tended to settle strongly on the opposing side, opposing the idea of sophisticated innate language modules with a strictly immutable idea of grammatical acceptability implemented in a physical Language Aquisition Device ``as real as the heart or the liver" (Chomsky in Piatelli-Palmarini, M.,1979).

Although Berwick (2016b) asks ``Why take a chance?" is it Berwick and Chomsky who are relying on chance in saying language is innate rather than learned, and relegating the problem to mysterious and unknowable processes of natural selection and evolution, rather than allowing for learning by far more powerful members of our computational learning arsenal?

\subsection*{``Colorless green ideas sleep furiously''}

Ironically, Chomsky's grammatical but supposedly nonsensical sentence, reflected in Berwick et al.\ (2016a)'s title and 120 permutation corpus, are used by their opponents to characterize this ``settled" debate. They may think they have settled their baby to sleep, but many remain unimpressed by the unripe ideas of Chomsky (1957-1995), and their lack of ability to paint a realistic world picture that encompasses the rich multimodal technicolor of \em language \em and \em ontology \em and \em understanding \em and \em communication \em and \em interaction\em... within our \em world\em, our \em society \em and our \em culture\em . Perhaps we have let sleeping ideas lie too long. The dreams that we pursue so furiously as Linguistic and Psycholinguistic researchers are still shaped, one way or another, by the assumptive embers of this sleepy smouldering debate. 

\nocite{*}
\bibliography{cacl2016}
\bibliographystyle{acl2016}
\parindent=0em
\everypar{\small\hangindent=1em}
\end{document}